\documentclass[10pt,journal]{IEEEtran}

\usepackage[T1]{fontenc}
\usepackage[table]{xcolor}
\usepackage{cite}
\usepackage{amsmath,amssymb,bm}
\usepackage{graphicx}
\usepackage{booktabs,array,multirow}
\usepackage{algpseudocode}
\usepackage{xcolor}
\usepackage[hidelinks]{hyperref}
\usepackage[capitalise,nameinlink,noabbrev]{cleveref}
\usepackage{algorithm}
\graphicspath{{figure/}}
\DeclareGraphicsExtensions{.pdf,.png,.jpg,.jpeg}
\newif\ifuseoriginalfigures
\useoriginalfiguresfalse

\title{MARS: Dynamic Routing of Reinforcement-Learned Atomic Skills for Multi-Stage Robotic Manipulation}
\author{Zequan Zeng and Kai Wang
\thanks{Kai Wang is the corresponding author.}%
}

\author{%
Zequan Zeng\textsuperscript{1},
Junqin Yang\textsuperscript{1},
Kai Wang\textsuperscript{1,*}
\thanks{\textsuperscript{1}School of Automation, Chongqing University, Chongqing, 400044, China.}%

\thanks{\textsuperscript{*}Corresponding author: Kai Wang.}%
}

\begin{document}
\maketitle
\begin{abstract}
Vision-Language-Action (VLA) models have demonstrated strong capabilities in multi-task robotic manipulation. However, they still face limitations in control precision and execution stability during critical manipulation stages, such as object grasping, alignment and insertion. To combine general-purpose manipulation capabilities with task-specific fine-grained control, while enabling effective coordination between generalist and specialized policies throughout multi-stage manipulation tasks, this paper proposes MARS (Modular Atomic-skill Routing System), a dynamic routing framework for reinforcement learning-based atomic skills. MARS integrates a frozen generalist VLA policy with multiple independently trained atomic skill policies. Using a shared RL Token state representation and the reference action of VLA, specialized control policies are optimized through reinforcement learning for different cr`itical manipulation stages. For high-level policy selection, we develop a dynamic router that integrates temporal context and skill semantics. The router employs Mamba-based recurrent encoding and a historical feature window to capture state evolution during manipulation. Skill descriptions of candidate policies are encoded into semantic queries, and attention mechanisms are used to establish correspondences between skill semantics and temporal state features. Experimental results show that MARS achieves an average Macro-F1 score of 90.47\% across four LIBERO task suites under fixed-trajectory evaluation, outperforming an MLP-based router by 7.08 percentage points. In the simulated Tube Transport task, the full-task success rate increases from 20.04\% to 85.67\%. These results demonstrate that stage-specific reinforcement learning and temporal state-aware dynamic policy routing can improve multi-policy coordination and overall task execution performance in multi-stage robotic manipulation.
\end{abstract}


\section{Introduction}
\label{sec:intro}

\begin{figure*}[t]
  \centering
  \includegraphics[width=\textwidth]{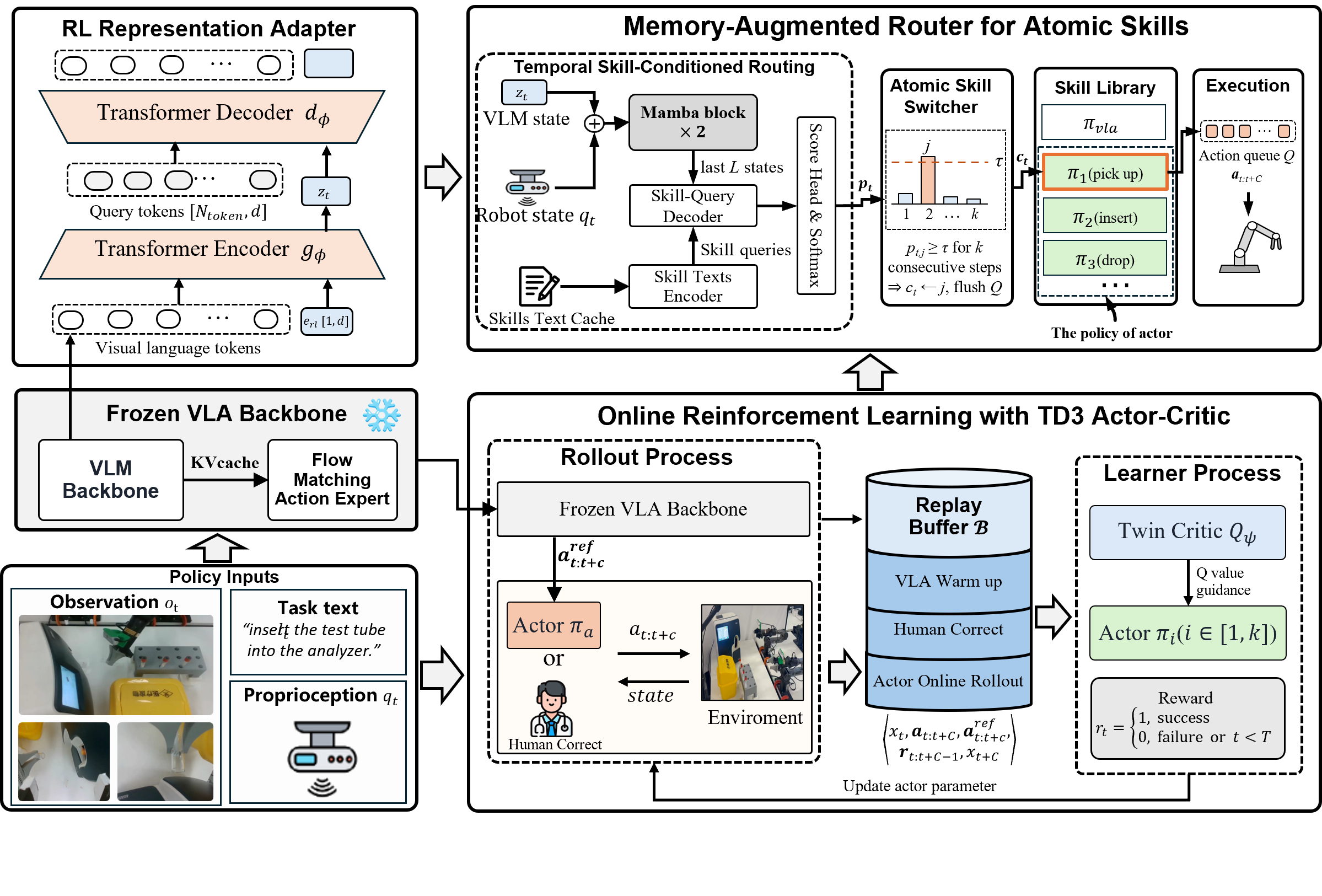}
  \caption{\textbf{Overview of the MARS framework.} MARS consists of three main components: shared state feature extraction, atomic skill policy learning, and dynamic skill routing. A shared RL Token adapter extracts compact state representations from the frozen VLA, supporting reinforcement learning-based atomic skill optimization with VLA reference actions. The dynamic router then combines historical state information and skill semantics to coordinate policy selection and switching during multi-stage manipulation.}
  \label{fig:framework}
\end{figure*}
Robotic manipulation tasks, such as multi-object pick-and-place, object transfer, and laboratory sample handling, typically involve a sequence of operations, including grasping, transportation, alignment, placement, and insertion. These stages impose different requirements on the underlying control policy. General-purpose policies are often sufficient for relatively simple operations, such as object transportation and robot resetting, whereas alignment and insertion require higher control precision and execution stability. Consequently, relying on a single control policy throughout an entire manipulation task may make it difficult to simultaneously achieve general-purpose manipulation capabilities and precise control during critical stages. Combining pretrained generalist policies with skill-specific control policies optimized for particular manipulation requirements provides a promising approach to improving multi-stage robotic manipulation. However, different policies have distinct operating conditions and applicable ranges. Selecting an appropriate policy according to the robot state and task progress, while ensuring smooth transitions between policies, remains an important challenge in multi-policy coordination~\cite{huang2026chembot,pang2026scivla}.

In recent years, generalist robotic policies, including RT-1, RT-2, Octo, OpenVLA, $\pi_0$, $\pi_{0.5}$, and SmolVLA, have demonstrated promising multi-task manipulation capabilities through large-scale robot data pretraining and joint modeling of visual, language and action information~\cite{brohan2022rt1,brohan2023rt2,ghosh2024octo,openx2024oxe,kim2024openvla,black2024pi0,intelligence2025pi05,shukor2025smolvla}. Nevertheless, the performance of these generalist policies varies across manipulation stages, particularly in tasks involving precise alignment and contact-constrained operations, where action accuracy and execution stability remain challenging~\cite{luo2025hilserl,xu2026rlt}. To further improve robotic manipulation performance, recent studies have investigated reinforcement learning (RL)-based post-training methods for VLA policies. One line of research improves manipulation performance by optimizing the action generation process or updating selected model parameters, as demonstrated by VLA-RL, SimpleVLA-RL, iRe-VLA, and $\pi_{0.6}$~\cite{lu2025vlarl,li2025simplevlarl,guo2025irevla,intelligence2025recap}. Another line of research aims to preserve the core capabilities of pretrained models while enhancing task-specific control through lightweight control modules, residual policies, or independently trained RL policies. Representative methods include RL Token (RLT), PARTS, Object-Centric Residual RL, and VLaRL~\cite{xu2026rlt,su2026pretraining,kim2026object,saito2026vlarl}. These studies have demonstrated the effectiveness of reinforcement learning in improving robotic manipulation from different perspectives and have provided a foundation for integrating pretrained generalist policies with specialized control strategies.

Although reinforcement learning can improve the control performance of VLA models in specific manipulation tasks, optimizing a single policy for an entire multi-stage task remains challenging. When only a binary terminal reward indicating full-task success or failure is available, the policy receives no reward feedback for intermediate actions that contribute positively to task progress. This makes it difficult to accurately estimate the contribution of actions at different stages to the final task outcome, thereby complicating state-action value estimation and policy optimization. One possible solution is to decompose critical stages with clearly defined manipulation objectives into independent atomic skills and optimize each skill separately through reinforcement learning. By assigning local objectives and reward signals to individual atomic skills, specialized policies can be trained to address specific control requirements while reducing their dependence on full-task rewards. However, independently trained atomic skills only improve control performance within their respective stages. Determining when to activate each skill and how to coordinate transitions between specialized policies and the generalist VLA remains a key challenge for multi-policy coordination. Several studies have explored skill organization and dynamic routing mechanisms from different perspectives. AtomicVLA organizes manipulation capabilities through atomic skill modeling and an expert-based architecture, SMoDP introduces semantically structured policy specialization for compositional robotic manipulation, and SWAP enables dynamic selection among multiple VLA policies through policy value estimation~\cite{zhang2026atomicvla,deng2026smodp,das2026swap}.  These approaches offer different solutions for atomic skill modeling, policy specialization, and dynamic selection. Nevertheless, for a system consisting of a frozen generalist VLA and multiple independently trained RL-based atomic skill policies, determining which policy is most appropriate at different stages of task execution remains a key challenge. Since individual policies have different control objectives and operating ranges, atomic skill selection should not rely solely on predefined manipulation stages but should also account for state changes throughout the sequential execution process. Meanwhile, descriptions of candidate policies provide policy-specific semantic information that can help the router learn correspondences between candidate skills and temporal state contexts. Motivated by these considerations, we formulate dynamic policy routing from two complementary perspectives: temporal state modeling and candidate policy matching. Specifically, temporal encoding is used to extract contextual information from sequential robot states, while skill descriptions are transformed into conditioned semantic queries that interact with temporal state features. This enables the router to construct policy-specific matching representations and provides a basis for dynamically selecting between the generalist VLA and reinforcement-learned atomic skill experts.

To address these challenges, we propose MARS (Modular Atomic-skill Routing System), a dynamic routing framework for reinforcement learning-based atomic skills in multi-stage robotic manipulation. MARS decouples local skill optimization from high-level policy selection, enabling coordinated control between a generalist VLA and multiple specialized atomic skill policies. For atomic skill optimization, we first construct a shared RL Token adapter based on intermediate features extracted from the frozen VLA backbone. The adapter produces compact state representations that can be reused across different skill policies. Each atomic skill policy then takes the shared state representation and reference action chunks of the frozen VLA as inputs and is independently optimized through reinforcement learning for a specific local manipulation objective. For high-level policy selection, we develop a dynamic router that integrates temporal context and skill semantics. The router employs a Mamba-base recurrent encoder~\cite{gu2023mamba} to capture temporal information from robot states and maintains a fixed-length historical feature window to preserve recent state changes. Meanwhile, the skill descriptions of candidate policies are encoded into semantic queries, which are further refined through self-attention to capture inter-skill relationships. Cross-attention is then applied to extract skill-relevant information from historical state features, providing contextual representations for subsequent policy selection. The skill-conditioned features are then used to predict routing probabilities for the generalist VLA and each atomic skill policy.

The main contributions of this paper are summarized as follows:

\begin{enumerate}
    \item We propose MARS, a multi-policy coordination framework that integrates a frozen generalist VLA with independently optimized reinforcement learning-based atomic skill policies using shared RL Token representations and VLA reference actions.

    \item We develop a dynamic policy router that combines Mamba-based temporal modeling and skill-conditioned attention to select appropriate control policies based on temporal state information and candidate skill semantics.
    
     \item We evaluate MARS through simulation and real-world robotic experiments, demonstrating its effectiveness in atomic skill execution, dynamic policy routing, and full-task manipulation performance.
\end{enumerate}

\section{Method}
\label{sec:method}

Multi-stage robotic manipulation involves operations with different control requirements, requiring both general-purpose manipulation capabilities and specialized control during critical stages. To leverage the complementary strengths of generalist and specialized policies, we propose MARS, a multi-policy coordination framework that decouples local atomic skill optimization from high-level policy selection. In this framework, a frozen generalist VLA handles basic manipulation and provides reference actions for atomic skill learning, while multiple atomic skill policies undergo independent reinforcement learning optimization for specific manipulation objectives. A high-level dynamic router then selects appropriate control policies based on robot state information and candidate skill semantics, as illustrated in \cref{fig:framework}.

\subsection{Framework Overview and Problem Formulation}
\label{sec:overview}
MARS separates high-level policy selection from low-level action generation. The set of candidate low-level policies is defined as:
\begin{equation}
  \Pi=\{\pi_0,\pi_1,\ldots,\pi_K\}
  \label{eq:policy_set}
\end{equation}
where $\pi_0$ denotes the frozen generalist VLA policy, and $\pi_j$ ($j=1,\ldots,K$) represents an atomic skill policy trained for a specific manipulation objective. Each atomic skill policy generates actions through its Actor network $\mu_j$. In the following, $\pi_j$ refers to the complete candidate low-level policy, whereas $\mu_j$ denotes its corresponding action generation network.

At time step $t$, the robot receives an observation $o_t=(I_t,l,q_t)$, where $I_t$ is the visual observation, $l$ denotes the task instruction, and $q_t\in\mathbb R^{d_q}$ represents the normalized robot proprioceptive state. The visual-language features extracted by the frozen VLA are compressed into $z_t^{\mathrm{RL}}\in\mathbb R^{d_z}$ through the shared RL Token encoder (see \cref{sec:rltoken}). The resulting feature is concatenated with the proprioceptive state to construct the state representation used by the atomic skill policies:
\begin{equation}
  s_t=[z_t^{\mathrm{RL}};q_t]\in\mathbb R^{d_z+d_q}
  \label{eq:state}
\end{equation}
Let $s_{1:t}=(s_1,\ldots,s_t)$ denote the state sequence up to time step $t$, and let $\mathcal{S}=\{\ell_0,\ldots,\ell_K\}$ denote the set of skill descriptions associated with the candidate policies. The router $R_{\phi}$ predicts a probability distribution over the candidate policies:
\begin{equation}
  p_t=R_\phi(s_{1:t},\mathcal{S}),\quad
  p_t\in[0,1]^{K+1},\quad
  \sum_{j=0}^K p_{t,j}=1
  \label{eq:router}
\end{equation}
where $p_{t,j}$ represents the predicted probability of selecting policy $\pi_j$, conditioned on the state history and the skill descriptions of the candidate policies.

\subsection{Shared RL Token Feature Extraction}
\label{sec:rltoken}
To avoid redundant learning of visual-language representations across different atomic skill policies, we construct a shared lightweight RL Token adapter based on the feature compression approach introduced in RLT~\cite{xu2026rlt}. The adapter learns a compact state representation by reconstructing intermediate VLA features. Once trained, its encoder is frozen and shared by the atomic skill policies and the dynamic router.

Given the visual observation $I_t$ and task instruction $l$ at time step $t$, the frozen VLA backbone $F_{\mathrm{VLA}}$ produces a sequence of multimodal features:
\begin{equation}
  H_t=F_{\mathrm{VLA}}(I_t,l)
  \in\mathbb R^{N_t\times d_h}
  \label{eq:vla_hidden}
\end{equation}
where $N_t$ denotes the token sequence length and $d_h$ is the feature dimension. The valid tokens are first normalized along the feature dimension to obtain $\bar H_t$, and using a binary mask $m_t\in\{0,1\}^{N_t}$ to indicate valid token positions. Positional embeddings and a learnable aggregation token are then incorporated into the feature sequence, which is processed by a Transformer encoder $g_\psi$ to obtain a compact, fixed-dimensional representation:
\begin{equation}
  z_t^{\mathrm{RL}}
  =g_\psi(\bar H_t,m_t)
  \in\mathbb R^{d_z}
  \label{eq:rl_token}
\end{equation}
To preserve relevant information from the intermediate VLA features, a decoder $g_\eta$ reconstructs the feature sequence from the compact representation $z_t^{\mathrm{RL}}$, yielding $\hat H_t=g_\eta(z_t^{\mathrm{RL}})$. The reconstruction loss is then computed as the mean squared error between the reconstructed features $\hat H_t$ and the normalized features $\bar H_t$ over valid token positions:
\begin{equation}
  \mathcal L_{\mathrm{rec}}
  =\frac{
    \sum_{n=1}^{N_t}m_{t,n}
    \left\|\hat H_{t,n}-\bar H_{t,n}\right\|_2^2
  }{
    d_h\max\left(1,\sum_{n=1}^{N_t}m_{t,n}\right)
  }
  \label{eq:reconstruction}
\end{equation}
After training, the decoder is discarded and the encoder parameters are frozen, allowing the atomic skill policies and the dynamic router to share a unified state representation.

\label{sec:mars}
\begin{figure}[t]
  \centering
  \includegraphics[width=0.5\textwidth]{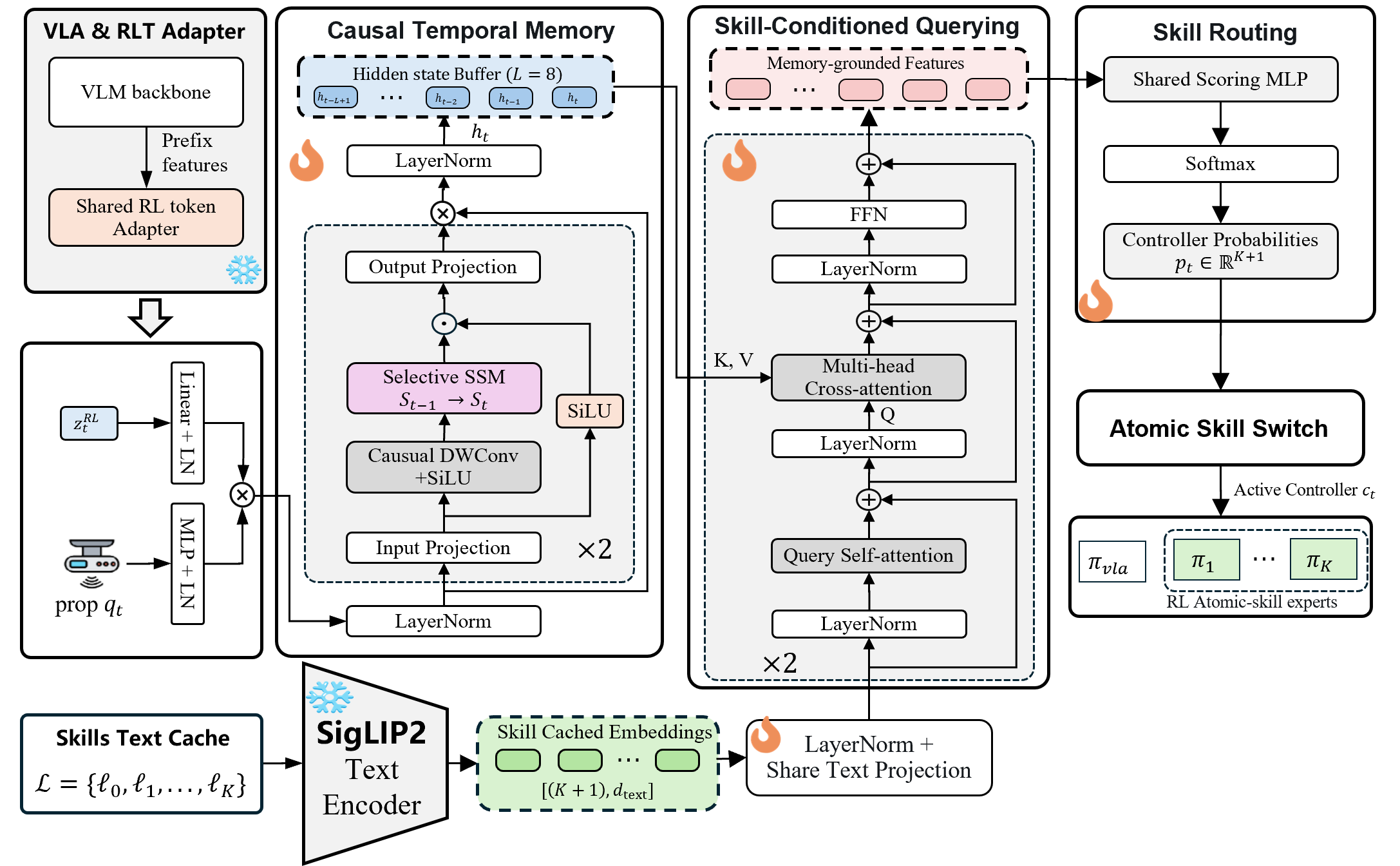}
  \caption{\textbf{Architecture of the dynamic skill router.} The dynamic skill router combines Mamba-based temporal encoding with a historical feature window to capture temporal context. Candidate skill descriptions form semantic queries, which interact through self-attention and extract skill-relevant historical features via cross-attention. A shared scoring network predicts routing probabilities, which guide the subsequent policy switching mechanism.}
  \label{fig:router}
\end{figure}

\subsection{Atomic Skill Policy Learning with Shared RL Token Representations and Reference Action Constraints}
\label{sec:experts}
To improve control performance during critical manipulation stages, we train specialized atomic skill policies for different local manipulation objectives using reinforcement learning. Each policy is independently optimized using the Twin Delayed Deep Deterministic Policy Gradient (TD3) algorithm~\cite{fujimoto2018td3}, with actions predicted by the VLA backbone serving as references. This design allows the skill policies to refine their control behaviors for specific manipulation requirements while leveraging the action prior provided by the pretrained VLA.

During atomic skill training, both the VLA backbone and the shared RL Token encoder remain frozen, and only the Actor and Critic networks of the corresponding skill policy are updated. For the $j$-th atomic skill policy, let $\mu_j$ denote its Actor network and $Q_{j,1}$ and $Q_{j,2}$ denote its two Critic networks. The corresponding target networks are denoted by $\bar{\mu}_j$, $\bar{Q}_{j,1}$, and $\bar{Q}_{j,2}$, respectively.

\subsubsection{Action Generation and Replay Sample Construction}
At control time step $t$, the VLA backbone $\pi_0$ generates an action sequence of length $H$ conditioned on the current observation $o_t$. The first $C$ actions are extracted to form the reference action chunk:
\begin{align}
A_t^{\mathrm{VLA}} &\sim \pi_0(\cdot\mid o_t),\\
A_t^{\mathrm{ref}} &= A_t^{\mathrm{VLA}}[0:C],
\quad A_t^{\mathrm{ref}}\in\mathbb R^{C\times d_a}
\end{align}
where $d_a$ denotes the dimension of a single action. Given the shared state representation $s_t$ and the reference action chunk $A_t^{\mathrm{ref}}$, the Actor $\mu_j$ generates an action chunk of length $C$:
\begin{equation}
A_t^j=\mu_j(s_t,A_t^{\mathrm{ref}})
\in\mathbb R^{C\times d_a}
\end{equation}
The reference action chunk serves as an additional input to the Actor and is also used to construct a reference action constraint during training. In this way, the atomic skill policy can adjust its control actions according to local manipulation objectives while retaining the action prior of the VLA backbone.

We adopt an off-policy reinforcement learning approach and maintain a separate replay buffer $\mathcal D_j$ for each atomic skill policy. The replay buffers contain warm-up trajectories collected using the VLA backbone, online interaction trajectories generated by the atomic skill policies, and trajectories involving human interventions. To improve data utilization, we construct replay samples by extracting overlapping trajectory segments through a sliding-window procedure. Specifically, segments of length $C$ are sampled from continuous trajectories with a stride of two time steps, forming the following training samples:
\begin{equation}
    \begin{aligned}
    \big(
    &s_t,\,
    A_t^{\mathrm{exec}},\,
    r_{t:t+C-1},\,
    s_{t+C},\\
    &A_t^{\mathrm{ref}},\,
    A_{t+C}^{\mathrm{ref}},\,
    \delta_{t+C}
     \big)
    \end{aligned}
\end{equation}
where $A_t^{\mathrm{exec}}$ denotes the action chunk executed over the sampled trajectory segment, $r_{t:t+C-1}$ is the corresponding reward sequence, and $s_{t+C}$ represents the state at the end of the segment. The binary variable $\delta_{t+C}\in{0,1}$ denotes the termination flag of the current atomic skill, which is set to $1$ when the local manipulation objective is achieved or the active policy is switched to another policy.

Reference actions are aligned with individual time steps within each trajectory segment and may therefore originate from VLA predictions generated at different time steps. When human intervention occurs, the corresponding intervention actions are used as references. By extracting overlapping segments from collected trajectories, this sampling strategy increases the number of available training samples and improves data utilization.

\subsubsection{TD3-based Atomic Skill Policy Optimization}
Each atomic skill policy is optimized using TD3, which employs twin Critic networks for action-value estimation, together with delayed policy updates and target policy smoothing to improve training stability. For a trajectory segment of length $C$ sampled from the replay buffer $\mathcal D_j$, the cumulative discounted reward is computed as:
\begin{equation}
    R_t=\sum_{i=0}^{C-1}\gamma^i r_{t+i}
\end{equation}
where $\gamma\in[0,1]$ denotes the discount factor. To construct the temporal-difference (TD) target, the target Actor $\bar{\mu}_j$ generates an action chunk conditioned on the state $s_{t+C}$ at the end of the segment and its corresponding reference action chunk $A_{t+C}^{\mathrm{ref}}$:
    \begin{equation}
    \widetilde A_{t+C}
    =\bar{\mu}_j(s_{t+C},A_{t+C}^{\mathrm{ref}})
    +\epsilon
\end{equation}
where $\epsilon$ denotes the target policy smoothing noise. To reduce overestimation bias, we use the smaller of the two target Critic estimates to construct the TD target:
\begin{equation}
  \begin{aligned}
  Y_t=\operatorname{clip}\Big(
      &R_t+\gamma^C(1-\delta_{t+C})\\
      &\quad\cdot\min_{i=1,2}
      \bar Q_{j,i}(s_{t+C},\widetilde A_{t+C}),
      Y_{\min},Y_{\max}
  \Big)
  \end{aligned}
\end{equation}
where $Y_{\min}$ and $Y_{\max}$ denote the lower and upper clipping bounds of the TD target, respectively. When $\delta_{t+C}=1$, the current atomic skill is considered terminated, and the value estimate at the next state is excluded from the TD target.

The two online Critic networks estimate the action values of the executed action chunk $A_t^{\mathrm{exec}}$ under the initial state $s_t$. Their parameters are updated by minimizing the squared temporal-difference errors:
\begin{equation}
  \mathcal L_{\mathrm{critic}}
  =\mathbb E_{\mathcal D_j}\left[
  \sum_{i=1}^{2}
  \left(
  Q_{j,i}(s_t,A_t^{\mathrm{exec}})-Y_t
  \right)^2
  \right]
\end{equation}

The Actor is optimized by combining action-value maximization with a reference action constraint. To prevent the Actor from relying solely on the VLA reference actions and simply reproducing them, we randomly replace the reference action chunk provided as input to the Actor with zeros with a probability of $50\%$ during training. Let $\widetilde A_t^{\mathrm{ref}}$ denote the modified reference input and $\widehat A_t^j=\mu_j(s_t,\widetilde A_t^{\mathrm{ref}})$ denote the corresponding Actor output. The Actor loss is defined as:
\begin{equation}
  \mathcal L_{\mathrm{actor}}
  =\mathbb E_{\mathcal D_j}\left[
  -\lambda_Q Q_{j,1}(s_t,\widehat A_t^j)
  +\lambda_{\mathrm{ref}}
  \big\|
  \widehat A_t^j-A_t^{\mathrm{ref}}
  \big\|_2^2
  \right]
  \label{eq:actor_loss}
\end{equation}
where $\lambda_Q$ and $\lambda_{\mathrm{ref}}$ are fixed weighting coefficients for the action-value term and the reference action constraint, respectively. The reference action constraint encourages the Actor to generate actions close to the VLA-predicted reference actions $A_t^{\mathrm{ref}}$, thereby reducing excessive deviations from the pretrained action prior during the early stages of online reinforcement learning.

\subsection{Dynamic Policy Routing with Temporal Context and Skill Semantics}
Through independent reinforcement learning optimization, individual atomic skill policies acquire specialized control capabilities for different manipulation objectives. However, completing multi-stage manipulation tasks requires effective coordination between the VLA backbone and these specialized policies. To achieve this, the system must select appropriate control policies based on the robot's current state and historical context throughout task execution. Accordingly, we develop a dynamic router that integrates temporal state information and candidate skill semantics to support adaptive policy selection, as illustrated in \cref{fig:router}. 

\subsubsection{Temporal Context Modeling}
\label{sec:fusion}
During sequential robotic manipulation, the suitability of a candidate policy depends not only on the current observation but also on information from preceding states. Therefore, we introduce a causal temporal modeling module into the routing network to incorporate historical context into policy selection. This module consists of a Mamba-based recurrent encoder and a fixed-length historical feature window. The former updates the temporal state representation, while the latter provides recent contextual features for subsequent skill-conditioned queries.

At time step $t$, we encode the shared RL Token feature $z_t^{\mathrm{RL}}$ and robot proprioceptive state $q_t$ using a linear layer and an MLP, respectively. We then concatenate the normalized features to obtain the state embedding $x_t$:
\begin{equation}
x_t=
\left[
\operatorname{LN}(\operatorname{Linear}(z_t^{\mathrm{RL}}));
\operatorname{LN}(\operatorname{MLP}_q(q_t))
\right],
\quad x_t\in\mathbb R^d
\end{equation}
where $d$ denotes the combined feature dimension. We feed $x_t$ into a causal temporal encoder comprising two Mamba blocks~\cite{gu2023mamba} to capture temporal dependencies in robot state sequences:
\begin{equation}
  (h_t,\xi_t)=\Phi_{\mathrm M}(x_t,\xi_{t-1}),
  \quad h_t\in\mathbb R^d
\end{equation}
where $\Phi_{\mathrm M}$ denotes the Mamba temporal encoder, $h_t$ represents the temporal feature at time step $t$, and $\xi_t$ denotes the recurrent cache, including the internal state-space variables and convolutional cache. Through recurrent updates, the encoder incorporates historical state information into the current representation without reprocessing the entire state sequence during online inference.

To retain recent temporal features, we maintain a historical feature window with a maximum length of $L$. Let $L_t=\min(t,L)$ denote the effective window length at time step $t$. We construct the historical feature window as:
\begin{equation}
\mathcal M_t=
  \left[
  h_{t-i}+e_i
  \right]_{i=0}^{L_t-1}
  \in\mathbb R^{L_t\times d}
\end{equation}
where $e_i\in\mathbb R^d$ denotes a learnable relative temporal embedding that encodes the temporal distance from the current time step. The window maintains the most recent $L$ temporal features in reverse chronological order and updates during task execution, providing historical context for subsequent skill-conditioned queries. By combining Mamba-based recurrent encoding with the historical feature window, the router captures temporal dependencies and retains recent contextual information to support dynamic policy selection.

\subsubsection{Skill Semantics-guided Policy Routing}
\label{sec:query}
Each candidate policy has distinct control capabilities, and its skill description provides semantic information for policy selection. To incorporate this information into dynamic routing, we encode candidate skill descriptions into semantic queries and use them to extract skill-relevant temporal features from the historical state representations.

Given the candidate policy set $\Pi=\{\pi_0,\ldots,\pi_K\}$, let $\ell_j$ denote the skill description of policy $\pi_j$, where $j=0,\ldots,K$. We employ a frozen SigLIP 2 text encoder (ViT-B/16)~\cite{tschannen2025siglip2} to extract semantic features, followed by $L_2$ normalization and a shared linear projection to obtain the skill query vector:
\begin{equation}
v_j=f_{\mathrm{skill}}(\ell_j)\in\mathbb R^d
\end{equation}
where $f_{\mathrm{skill}}$ denotes the semantic mapping module comprising text encoding, feature normalization, and linear projection. Since the skill descriptions remain fixed during task execution, we precompute and cache their query vectors to reduce online computational overhead.

To integrate skill semantics with temporal context, we construct a skill query module consisting of $N_q$ attention layers. Each layer first applies self-attention to capture semantic relationships among candidate skills. The updated skill queries then attend to the historical feature window $\mathcal M_t$ through cross-attention, using the skill queries as Queries and the historical features as Keys and Values. This process extracts skill-relevant temporal information and produces contextual representations for individual candidate policies:
\begin{equation}
\{f_{t,j}\}_{j=0}^{K}
=
\operatorname{Attn}
\left(
\{v_j\}_{j=0}^{K},\mathcal M_t
\right)
\end{equation}
Here, $\operatorname{Attn}$ denotes the multi-layer feature fusion module composed of inter-skill self-attention, cross-attention over historical features, and feed-forward networks. The feature $f_{t,j}\in\mathbb R^d$ represents the contextual representation of the $j$-th candidate policy conditioned on the current state history.

We then feed these representations into a shared scoring network $g_{\mathrm{score}}$ to compute policy matching scores and apply Softmax to obtain the routing probabilities:
\begin{equation}
\rho_{t,j}=g_{\mathrm{score}}(f_{t,j}),
\qquad
p_{t,j}=
\frac{\exp(\rho_{t,j})}
{\sum_{i=0}^{K}\exp(\rho_{t,i})}
\end{equation}
where $\rho_{t,j}$ and $p_{t,j}$ denote the matching score and routing probability of the $j$-th candidate policy, respectively. The routing probability distribution $p_t=[p_{t,0},\ldots,p_{t,K}]$ provides the basis for subsequent policy selection and switching.

We train the router through supervised learning. Since different skills may have varying execution durations, the training samples can exhibit class imbalance across candidate policies. To account for this imbalance, we adopt a class-frequency-weighted cross-entropy loss:
\begin{equation}
  \mathcal L_{\mathrm{route}}
  =-\frac{
  \sum_{t\in\mathcal T}w_{y_t}\log p_{t,y_t}
  }{
  \sum_{t\in\mathcal T}w_{y_t}
  },
  \qquad
  w_j=\frac{N}{(K+1)n_j}
\end{equation}
where $\mathcal T$ denotes the set of supervised time steps in the current training batch, $y_t\in\{0,\ldots,K\}$ represents the ground-truth policy label, $N$ is the total number of supervised samples in the training set, and $n_j$ denotes the number of samples belonging to the $j$-th policy class. Accordingly, $w_j$ represents the class weight assigned to policy $\pi_j$.

By integrating skill semantics with historical state information, the router learns to predict the suitability of candidate policies under different manipulation conditions, supporting dynamic policy selection throughout task execution.

\subsubsection{Dynamic Policy Switching}
\label{sec:switch}
The router predicts the routing probabilities of candidate policies based on historical state information and skill semantics. However, fluctuations in routing predictions across consecutive control steps may cause frequent policy switching and disrupt execution continuity. To address this issue, we introduce a switching mechanism that combines a probability threshold with consecutive confirmation to maintain stable policy execution.
At time step $t$, the router first identifies the candidate policy with the highest routing probability:
\begin{equation}
\hat c_t=
\arg\max_{j\in\{0,\ldots,K\}}p_{t,j}
\end{equation}
where $\hat c_t$ denotes the candidate policy index. The switching mechanism activates a new policy only when $\hat c_t$ differs from the current policy, its routing probability satisfies $p_{t,\hat c_t}\geq\tau$, and the same candidate meets these conditions for $k$ consecutive control steps. Otherwise, the system continues executing the current policy. Let $c_t$ denote the selected policy index, where $c_t=0$ corresponds to the VLA backbone and $c_t>0$ indicates the corresponding atomic skill policy.

\cref{alg:mars} presents the closed-loop execution procedure of MARS. At the beginning of each task, the system initializes the Mamba recurrent cache $\xi_0$, historical feature window $\mathcal M_0$, and policy switching module. At each control step, the router updates its temporal state representation and predicts the routing probabilities, while the switching module determines the active control policy.

To coordinate step-wise routing decisions with chunk-based action generation, we maintain an action queue $B$ that stores pending actions and a policy index $c_B$ that identifies their source. When the selected policy $c_t$ differs from $c_B$, the system clears the remaining queued actions to prevent interference between different policies. Whenever the queue becomes empty, the VLA backbone generates a new action chunk. The system either executes the VLA actions directly or uses them as references for the selected atomic skill policy to generate a specialized action chunk. This mechanism coordinates policy switching with action execution, enabling continuous closed-loop control during multi-stage manipulation.

\begin{algorithm}[htbp]
\caption{Closed-loop Execution and Policy Switching in MARS}
\label{alg:mars}
\begin{algorithmic}[1]
\Require VLA backbone $\pi_0$, atomic skill policies
$\{\mu_j\}_{j=1}^{K}$, router $R_\phi$,
switching module $S_{\tau,k}$,
maximum control steps $T_{\max}$

\State Initialize $Z_0=(\xi_0,\mathcal M_0)$
and $S_{\tau,k}$; set $c_0\gets 0$
\State $B\gets\varnothing,\quad c_B\gets 0$

\For{$t=1,\ldots,T_{\max}$ while the task is incomplete}
    \State Obtain observation $o_t$ and compute
    shared state $s_t$
    \State $(p_t,Z_t)\gets
    R_\phi.\mathrm{step}(s_t,Z_{t-1})$
    \State $c_t\gets S_{\tau,k}.\mathrm{update}(p_t)$

    \If{$c_t\neq c_B$}
        \State $B\gets\varnothing$
    \EndIf

    \If{$B=\varnothing$}
        \State $A_t^{\mathrm{VLA}}
        \sim\pi_0(\cdot\mid o_t)$
        \State $B\gets
        \begin{cases}
        A_t^{\mathrm{VLA}}, & c_t=0,\\
        \mu_{c_t}(s_t,A_t^{\mathrm{VLA}}[0:C]),
        & c_t>0
        \end{cases}$
        \State $c_B\gets c_t$
    \EndIf

    \State Pop and execute the first action $a_t$ from $B$
    \State Obtain the next observation $o_{t+1}$
\EndFor
\end{algorithmic}
\end{algorithm}
\section{Experiments}
\label{sec:experiments}
To evaluate the effectiveness of atomic skill optimization and dynamic policy routing in MARS, we conduct experiments on the LIBERO benchmark, a custom Tube Transport simulation environment, and a real robotic platform. These experiments assess local skill execution, skill routing performance, and full-task closed-loop execution, while further examining the applicability of MARS to real-world robotic manipulation. We organize our evaluation around four research questions:

\noindent\textbf{Q1: Does reinforcement learning improve the execution performance of individual atomic skills?}
We compare the success rates of the frozen VLA and corresponding atomic skill policies under identical stage-specific initial conditions to assess the benefits of optimizing local manipulation objectives.

\noindent\textbf{Q2: How do temporal context modeling and skill-semantic queries affect routing performance?}
We compare different routing architectures on identical fixed test trajectories and conduct component ablations to examine their effects on policy-label prediction.

\noindent\textbf{Q3: Can atomic skill policies and dynamic routing improve closed-loop full-task execution?}
We compare the frozen VLA, a monolithic full-task RL policy, and multi-policy systems with different routing configurations to evaluate their combined contributions.

\noindent\textbf{Q4: Can MARS achieve reliable closed-loop execution of the Tube Transport task on a real robot?}
We evaluate multi-tube sequential transfer on a physical robot using full-task success rate and sample completion rate, to assess the deployability of the proposed framework.

\subsection{Experimental Setup}
\label{sec:setup}

\subsubsection{Platforms and Task Configurations}
\label{sec:sim_tasks}
\begin{figure}[!t]
  \centering
  \includegraphics[width=\linewidth]{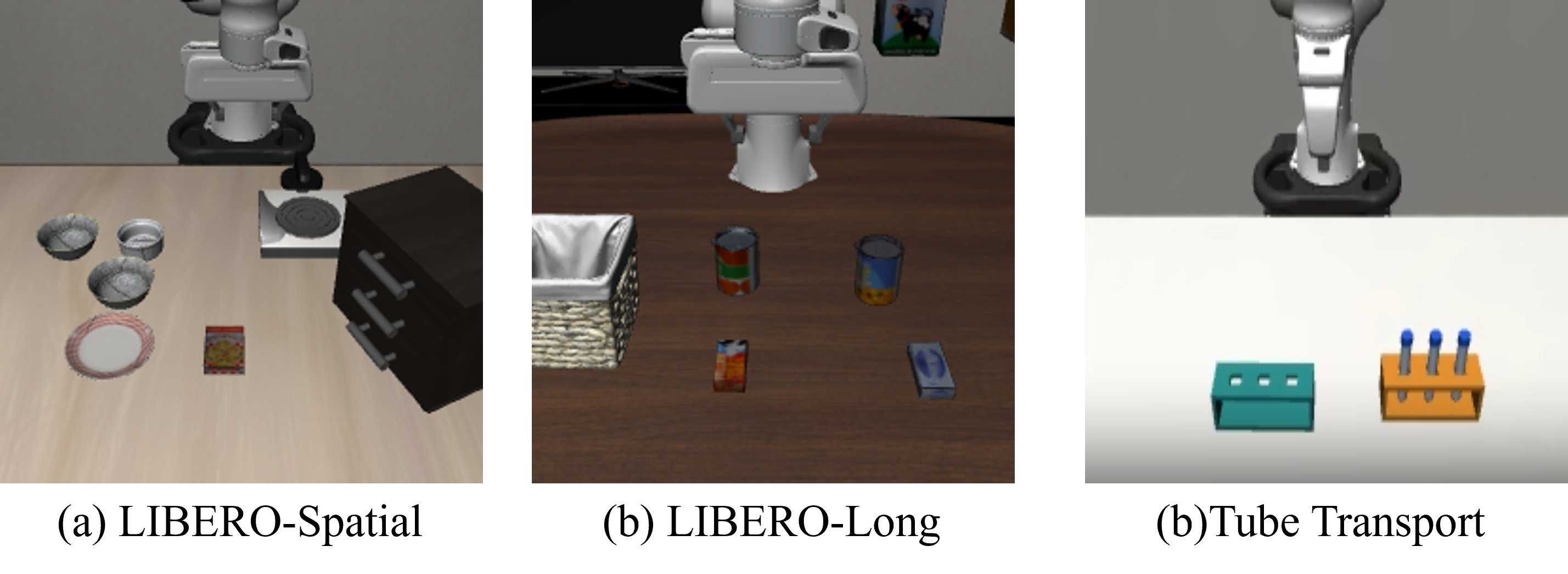}
  \caption{Simulation task environments.}
  \label{fig:tasks}
\end{figure}
We conduct experiments on the LIBERO benchmark, a custom Tube Transport simulation environment, and a physical robot platform to evaluate MARS under different manipulation conditions. We use a single NVIDIA RTX 4090 GPU with 24~GB of memory for model training and evaluation.

\noindent\textbf{Simulation tasks.}
LIBERO~\cite{liu2023libero}, built on robosuite~\cite{zhu2020robosuite}, consists of four task suites: Spatial, Object, Goal, and Long, covering diverse manipulation scenarios involving spatial relationships, objects, task goals, and sequential operations. We further develop a Tube Transport simulation environment in which the robot sequentially grasps three test tubes, transports them to designated slots in a target rack, and completes insertion before returning to its initial pose, as shown in Fig.~\ref{fig:tasks}.

\noindent\textbf{Real-robot platform.}
Our real-world experimental platform consists of an AgileX piper robotic arm equipped with a parallel gripper and an Intel RealSense D435i camera (RGB, $640\times480$ at 30~Hz). The real-robot task follows the same manipulation sequence as Tube Transport in simulation. With fixed source and target racks and camera placement, the robot sequentially grasps, transfers, aligns, inserts, and releases three test tubes in the specified order before returning to its initial pose.

\subsubsection{Baselines and Comparison Methods}
\label{sec:baselines}
To evaluate the effectiveness of MARS, we consider two groups of comparison methods: base policies and routing methods.

\noindent\textbf{Base policies.}
\emph{Frozen VLA} directly executes the entire task using the pretrained generalist policy without atomic skill policies or dynamic routing. \emph{Monolithic RLT} adopts the shared RL Token representation and VLA reference action constraints of RLT~\cite{xu2026rlt}, but trains a single RL Actor to optimize the entire manipulation task.

\noindent\textbf{Routing methods.}
We compare MARS with MLP, LSTM, and Transformer routers to evaluate the effects of different routing architectures. The MLP predicts policy labels from the current state, while the LSTM and Transformer incorporate temporal context through recurrent hidden states and temporal attention, respectively. We further introduce two ablation variants, \emph{MARS w/o Temporal Memory} and \emph{MARS w/o Semantic Queries}, to examine the contributions of temporal context modeling and skill-conditioned queries.

\subsubsection{Evaluation Metrics}
\label{sec:metrics}

\noindent\textbf{Success rate (SR).}
SR measures the percentage of trials that successfully complete the target operation within the specified control budget. We follow the official success criteria for LIBERO. For Tube Transport, a trial succeeds only when the robot transfers and inserts all three tubes in the specified order and returns to its initial pose. We also use SR to evaluate individual atomic skills under matched initial conditions.

\noindent\textbf{Sample completion rate (SCR).}
To measure partial task progress in Tube Transport, we calculate the proportion of successfully inserted and released tubes across evaluation episodes:
\begin{equation}
    \mathrm{SCR}
    =\frac{1}{|\mathcal E|}
    \sum_{e\in\mathcal E}
    \frac{n_e^{\mathrm{done}}}{n^\mathrm{total}_e}\times100\%
\end{equation}
where $\mathcal E$ denotes the evaluation episodes and $n_e^{\mathrm{done}}$ is the number of successfully completed tube insertions in episode $e$.

\noindent\textbf{Skill routing performance.}
We evaluate routing performance on fixed test trajectories using Macro-F1~\cite{sokolova2009systematic}, which averages the F1 scores across all candidate policy classes:
\begin{equation}
  \mathrm{Macro\mbox{-}F1}
  =\frac{1}{K+1}\sum_{k=0}^{K}
  \frac{2P_kR_k}{P_k+R_k}\times100\%
\end{equation}
where $P_k$ and $R_k$ denote the precision and recall of class $k$.

\subsubsection{Training and Evaluation Details}
\label{sec:impl}

We use a frozen SmolVLA as the generalist backbone and train separate TD3 atomic skill policies with shared RL Token representations. The router combines Mamba-based temporal modeling with skill-semantic queries. During router training, we freeze the VLA backbone, feature adapter, and atomic skill policies, and update only the routing network. Tables~\ref{tab:router_hyperparams} and~\ref{tab:rl_hyperparams} summarize the main hyperparameters.

For a consistent comparison, all methods use the same evaluation initial states. We conduct three independent evaluations with different random seeds and report the mean and standard deviation. For local skill evaluation, we use matched stage-specific initial conditions, while full-task experiments evaluate each method independently from identical initial states. We compare routing architectures and ablation variants on the same fixed test trajectories and select the routing models based on validation Macro-F1.

\begin{table}[t]
  \centering
  \caption{Main hyperparameters of the dynamic router.}
  \label{tab:router_hyperparams}
  \footnotesize
  \setlength{\tabcolsep}{4pt}
  \renewcommand{\arraystretch}{1.12}
  \begin{tabular}{@{}p{0.61\columnwidth}p{0.30\columnwidth}@{}}
    \toprule
    Hyperparameter & Value \\
    \midrule
    Feature dimension $d$ / attention heads & 128 / 4 \\
    RL Token / proprioceptive input dimensions & 960 / 8 \\
    Mamba blocks $N_b$ / convolution width $d_c$ & 2 / 4 \\
    State-space dimension $d_s$ & 16 \\
    History window $L$ / query layers $N_q$ & 8 / 2 \\
    Text encoder & SigLIP 2 ViT-B/16 \\
    Optimizer / learning rate & AdamW / $10^{-3}$ \\
    Training epochs / batch size & 30 / 8 \\
    Switching threshold $\tau$ / confirmations $k$ & 0.6 / 2 \\
    \bottomrule
  \end{tabular}
\end{table}

\begin{table}[t]
  \centering
  \caption{Main training hyperparameters for RL Token and TD3.}
  \label{tab:rl_hyperparams}
  \footnotesize
  \setlength{\tabcolsep}{4pt}
  \renewcommand{\arraystretch}{1.12}
  \begin{tabular}{@{}p{0.61\columnwidth}p{0.30\columnwidth}@{}}
    \toprule
    Hyperparameter & Value \\
    \midrule
    VLA backbone & SmolVLA \\
    RL Token / proprioceptive dimensions & 960 / 8 \\
    Adapter training objective & Intermediate-feature reconstruction \\
    RL algorithm / number of Critics & TD3 / 2 \\
    Actor / Critic hidden layers & $[256,256]$ / $[256,256]$ \\
    Action chunk length $C$ / sampling stride & 5 / 2 \\
    Actor / Critic learning rate & $3\times10^{-4}$ \\
    Reward type / reference-input dropout & Sparse / 0.5 \\
    \bottomrule
  \end{tabular}
\end{table}

\subsection{Main Result}
\subsubsection{Effectiveness of Atomic Skill Policies (Q1)}
\label{sec:results}
\label{sec:exp_local}
MARS preserves the generalist VLA's capabilities in routine manipulation while employing specialized atomic skill policies for precision-sensitive operations. To evaluate the effectiveness of these policies before examining their contributions to full-task performance, we compare their local execution success rates against those of the frozen VLA. Specifically, we select representative manipulation stages from LIBERO-Spatial and Tube Transport and evaluate both policies under identical stage-specific initial conditions. Table~\ref{tab:expert_local} presents the results.

As shown in Table~\ref{tab:expert_local}, the reinforcement learning-based atomic skill policies consistently outperform the frozen VLA across all four local operations. The improvement is particularly evident in precision tasks, with insertion success on Tube Transport increasing from 69.57\% to 96.85\%. These results demonstrate the benefits of targeted reinforcement learning optimization for local manipulation skills. However, these evaluations focus on individual skills under predefined initial conditions, without accounting for policy selection, switching, or error accumulation during full-task execution. Therefore, improved local skill performance does not necessarily guarantee higher full-task success rates. 
\subsubsection{Routing Performance and Ablation Studies (Q2)}
\label{sec:exp_routing}
\label{sec:exp_ablation}

While reinforcement learning-based atomic skill policies improve local manipulation performance, effective multi-policy execution also requires accurate policy selection. To evaluate the routing mechanism independently of low-level policy execution, we compare different routing architectures and component ablations on fixed LIBERO test trajectories with identical ground-truth policy labels. We use Macro-F1 to assess routing performance across the four LIBERO task suites.

As shown in Table~\ref{tab:routing}, MARS achieves the highest Macro-F1 across all four LIBERO task suites, with an average score of 90.47\%, exceeding the MLP baseline by 7.08 percentage points. The improvement is particularly notable on LIBERO-Long, where MARS achieves 89.85\% Macro-F1. Ablation results further show that removing temporal memory and skill-semantic queries reduces the average Macro-F1 to 84.97\% and 86.38\%, respectively. These results suggest that Mamba-based temporal encoding and the historical feature window help capture temporal context, while skill-semantic queries extract candidate-specific information to support more accurate policy routing. Together, these components contribute to improved routing performance, particularly in long-horizon manipulation tasks.

\begin{figure}[!t]
  \centering
  \includegraphics[width=\linewidth]{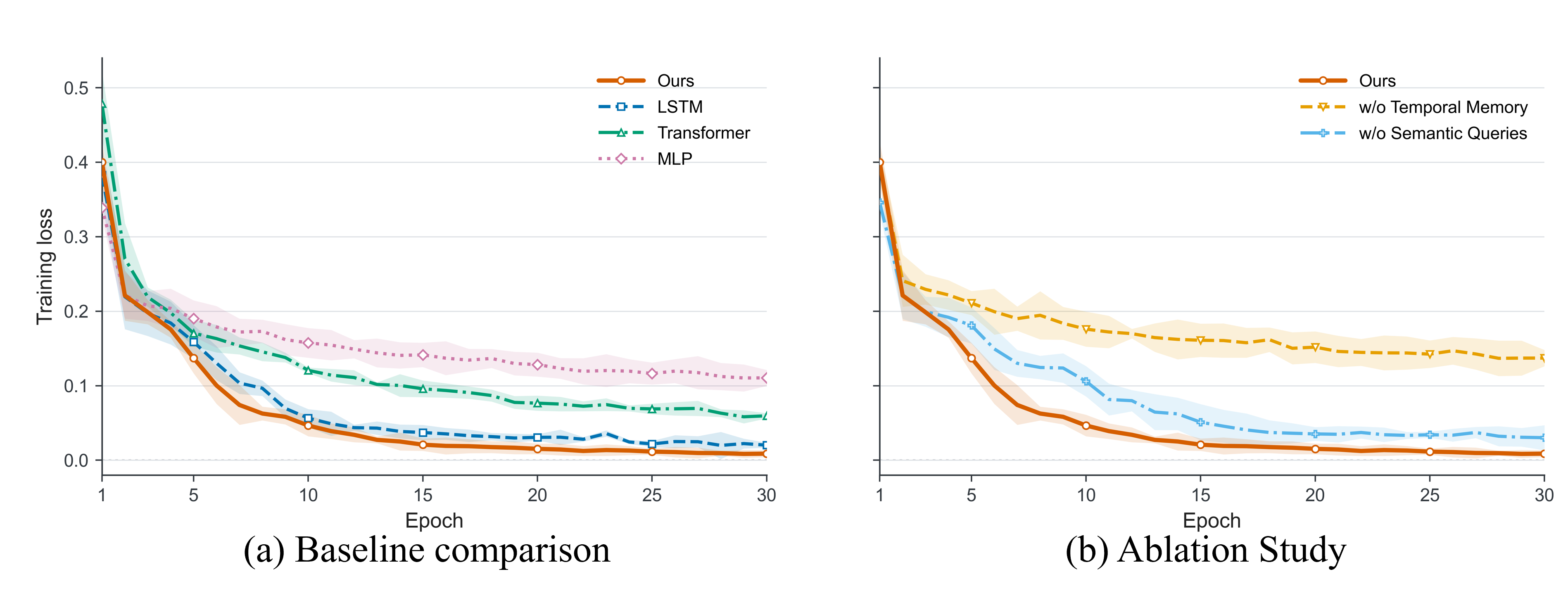}
  \caption{Training loss under different routing architectures and component ablations.}
  \label{fig:route_loss}
\end{figure}

Figure~\ref{fig:route_loss} presents the training loss curves for different routing configurations. As shown in Fig.~\ref{fig:route_loss}(a), all methods exhibit decreasing losses during training, while MARS maintains a comparatively low loss in later epochs. Similarly, Fig.~\ref{fig:route_loss}(b) shows higher training losses when either temporal memory or semantic queries are removed. These observations are consistent with the Macro-F1 results and provide additional evidence that both components improve the fit to the supervised routing objective. Overall, the results demonstrate that combining temporal context modeling with skill-semantic queries improves policy label prediction across diverse manipulation tasks, providing a basis for the closed-loop multi-policy evaluation in Q3.
\begin{table*}[!t]
  \centering
  \caption{Closed-loop performance in simulation. SR and SCR are reported in \% (mean $\pm$ standard deviation); Step denotes the reported execution-step metric.}
  \label{tab:main_sim}
  \scriptsize
  \setlength{\tabcolsep}{2pt}
  \renewcommand{\arraystretch}{1.16}
  \resizebox{\textwidth}{!}{%
  \begin{tabular*}{\textwidth}{@{\extracolsep{\fill}}lccccccc@{}}
    \toprule
    & \multicolumn{4}{c}{LIBERO SR $\uparrow$} & \multicolumn{3}{c}{Tube Transport} \\
    \cmidrule(lr){2-5}\cmidrule(l){6-8}
    Method & Spatial & Object & Goal & Long & SR $\uparrow$ & SCR $\uparrow$ & Step \\
    \midrule
    \multicolumn{8}{@{}l}{\emph{Base policies}} \\
    Frozen VLA & $69.13\pm4.58$ & $73.68\pm4.87$ & $66.22\pm3.72$ & $47.54\pm5.20$ & $20.04\pm4.06$ & $40.33\pm5.87$ & 912 \\
    Monolithic RLT & $84.12\pm3.61$ & $87.21\pm4.04$ & $81.24\pm2.13$ & $62.35\pm4.83$ & $43.97\pm5.07$ & $64.33\pm5.42$ & 821 \\
    \addlinespace[3pt]
    \multicolumn{8}{@{}l}{\emph{Routing baselines and oracle}} \\
    MARS w/o Temporal Memory & $86.89\pm3.20$ & $89.91\pm6.46$ & $83.73\pm2.89$ & $73.91\pm4.16$ & $74.52\pm5.56$ & $84.23\pm4.79$ & 725 \\
    MARS w/o Semantic Queries & $88.24\pm2.95$ & $90.83\pm1.95$ & $85.28\pm2.03$ & $76.66\pm3.80$ & $78.56\pm5.01$ & $86.89\pm4.34$ & 732 \\
    Rule-based Oracle Router & $\mathbf{93.33\pm3.06}$ & $92.15\pm3.31$ & $\mathbf{91.84\pm2.65}$ & $\mathbf{84.11\pm3.57}$ & $84.33\pm8.50$ & $87.44\pm3.60$ & 712 \\
    \midrule
    \rowcolor{black!5}
    \textbf{MARS (full)} & $91.33\pm1.78$ & $\mathbf{94.78\pm3.84}$ & $89.36\pm3.84$ & $81.54\pm3.22$ & $\mathbf{85.67\pm4.73}$ & $\mathbf{91.11\pm3.97}$ & 719 \\
    \bottomrule
  \end{tabular*}
  }
\end{table*}

\begin{table}[!t]
  \centering
  \caption{Local atomic skill performance compared with the frozen VLA. SR is reported in \%; $\Delta\mathrm{SR}$ in percentage points. Best SR values appear in bold.}
  \label{tab:expert_local}
  \footnotesize
  \setlength{\tabcolsep}{3pt}
  \renewcommand{\arraystretch}{1.15}
  \begin{tabular}{@{}llccc@{}}
    \toprule
    Task & Skill & Frozen VLA & Atomic skill & $\Delta\mathrm{SR}\uparrow$ \\
    \midrule
    \multirow{2}{*}{LIBERO-Spatial}
      & Grasp & $74.47\pm2.47$ & $\mathbf{94.61\pm2.13}$ & $+20.14$ \\
      & Place & $90.91\pm3.96$ & $\mathbf{99.12\pm0.35}$ & $+8.21$ \\
    \midrule
    \multirow{2}{*}{Tube Transport}
      & Grasp & $84.13\pm0.98$ & $\mathbf{99.33\pm0.58}$ & $+15.20$ \\
      & Insert & $69.57\pm2.15$ & $\mathbf{96.85\pm1.09}$ & $+27.28$ \\
    \bottomrule
  \end{tabular}
\end{table}

\begin{table}[t]
  \centering
  \caption{Skill routing performance on LIBERO (Macro-F1, \%).}
  \label{tab:routing}
  \scriptsize
  \setlength{\tabcolsep}{2.5pt}
  \renewcommand{\arraystretch}{1.15}
  \begin{tabular*}{\columnwidth}{@{\extracolsep{\fill}}lccccc@{}}
    \toprule
    Method & Spatial & Object & Goal & Long & Avg. \\
    \midrule
    \multicolumn{6}{@{}l}{\emph{Routing architectures}} \\
    MLP & 85.43 & 87.52 & 83.41 & 77.21 & 83.39 \\
    LSTM & 87.26 & 89.46 & \textbf{91.72} & 83.52 & 87.99 \\
    Transformer & \textbf{92.89} & 88.13 & 86.78 & 84.91 & 88.18 \\
    \addlinespace[2pt]
    \multicolumn{6}{@{}l}{\emph{Component ablations}} \\
    MARS w/o Temporal Memory & 85.85 & 88.05 & 83.79 & 82.19 & 84.97 \\
    MARS w/o Semantic Queries & 87.35 & 89.53 & 85.31 & 83.32 & 86.38 \\
    \midrule
    \rowcolor{black!1}
    \textbf{MARS (full)} & 91.63 & \textbf{92.12} & 88.27 & \textbf{89.85} & \textbf{90.47} \\
    \bottomrule
  \end{tabular*}
\end{table}

\begin{table}[!t]
  \centering
  \caption{Real-robot Tube Transport performance. Grasp and Insert report successful attempts/total attempts. Time denotes mean completion time over successful episodes.}
  \label{tab:real_robot_main}
  \scriptsize
  \setlength{\tabcolsep}{3pt}
  \renewcommand{\arraystretch}{1.18}
  \begin{tabular*}{\columnwidth}{@{\extracolsep{\fill}}lccccc@{}}
    \toprule
    Method & Grasp & Insert & SR (\%) $\uparrow$ & SCR (\%) $\uparrow$ & Time (s) $\downarrow$ \\
    \midrule
    Frozen VLA & 66/94 & 41/66 & 12.00 & 27.33 & 39.53 \\
    \textbf{MARS (full)} & 136/139 & 125/136 & 74.00 & 83.33 & 31.96 \\
    \bottomrule
  \end{tabular*}
\end{table}

\subsubsection{Multi-Policy Coordination and Full-Task Performance (Q3)}
\label{sec:exp_main}
The preceding experiments demonstrate the effectiveness of atomic skill policies in local manipulation and the router's ability to predict policy labels on fixed trajectories. However, these results do not directly reflect closed-loop performance, where policy decisions influence subsequent robot states. To evaluate multi-policy coordination during complete task execution, we integrate the dynamic router with the frozen VLA backbone and atomic skill policies, and conduct closed-loop experiments on LIBERO and Tube Transport.

Table~\ref{tab:main_sim} presents the full-task results. MARS achieves an average success rate of 89.25\% across the four LIBERO task suites, exceeding Frozen VLA and Monolithic RLT by 25.11 and 10.52 percentage points, respectively. On Tube Transport, MARS achieves 85.67\% SR and 91.11\% SCR, substantially outperforming both base policies. These improvements suggest that combining specialized atomic skills with dynamic routing benefits multi-stage manipulation. Compared with a single policy trained for the entire task, optimizing atomic skills for local objectives may also alleviate the difficulty of long-horizon reinforcement learning by shortening the reward propagation horizon. To further examine the contribution of dynamic routing, we evaluate different routing configurations while keeping the VLA backbone and atomic skill policies unchanged. Removing temporal memory and skill-semantic queries reduces the average LIBERO success rate by 5.64 and 4.00 percentage points, respectively. Similar performance reductions occur on Tube Transport, where both ablation variants achieve lower success rates than full MARS. These findings are consistent with the routing prediction results in Q2, suggesting that temporal context modeling and skill-semantic queries contribute to both policy selection and closed-loop task execution.

We additionally compare MARS with a Rule-based Oracle Router that switches policies using ground-truth manipulation-stage labels from the simulation environment. MARS achieves competitive performance without relying on such labels, outperforming the oracle on LIBERO-Object and Tube Transport while remaining below it on the other three LIBERO task suites. This comparison suggests that learned routing can coordinate atomic skills effectively without predefined stage-based switching decisions.

Overall, these results indicate that reinforcement learning-based atomic skills improve local control, while dynamic routing enables their coordination during complete task execution, contributing to stronger performance in multi-stage manipulation.

\subsection{Real-Robot Tube Transport Experiments (Q4)}
\label{sec:real_robot}

The previous experiments evaluate local atomic skills, routing prediction, and full-task closed-loop execution in simulation. Real-robot manipulation introduces additional challenges, including visual observation errors, robot control inaccuracies, and uncertain physical contacts. We therefore evaluate MARS on the AgileX PiPER platform described in Section~\ref{sec:sim_tasks} to examine whether the simulated improvements transfer to physical execution.

Table~\ref{tab:real_robot_main} summarizes the real-robot results. MARS raises full-task SR from 12.00\% with Frozen VLA to 74.00\%, an improvement of 62.00 percentage points, and increases SCR from 27.33\% to 83.33\%, a gain of 56.00 percentage points. MARS also achieves higher observed success ratios during grasping (136/139 versus 66/94) and insertion (125/136 versus 41/66). Among successful episodes, its mean completion time is 31.96~s, compared with 39.53~s for Frozen VLA.

These findings indicate that the specialized skills and routing mechanism remain effective in this physical Tube Transport setup. Together with the simulation results in Q3, they provide evidence of the framework's transferability to the tested robot configuration and manipulation task. Further evaluation under varying real-world layouts and operating conditions would be needed to establish broader deployment robustness.

\begin{figure}[t]
  \centering
  \includegraphics[width=0.5\textwidth]{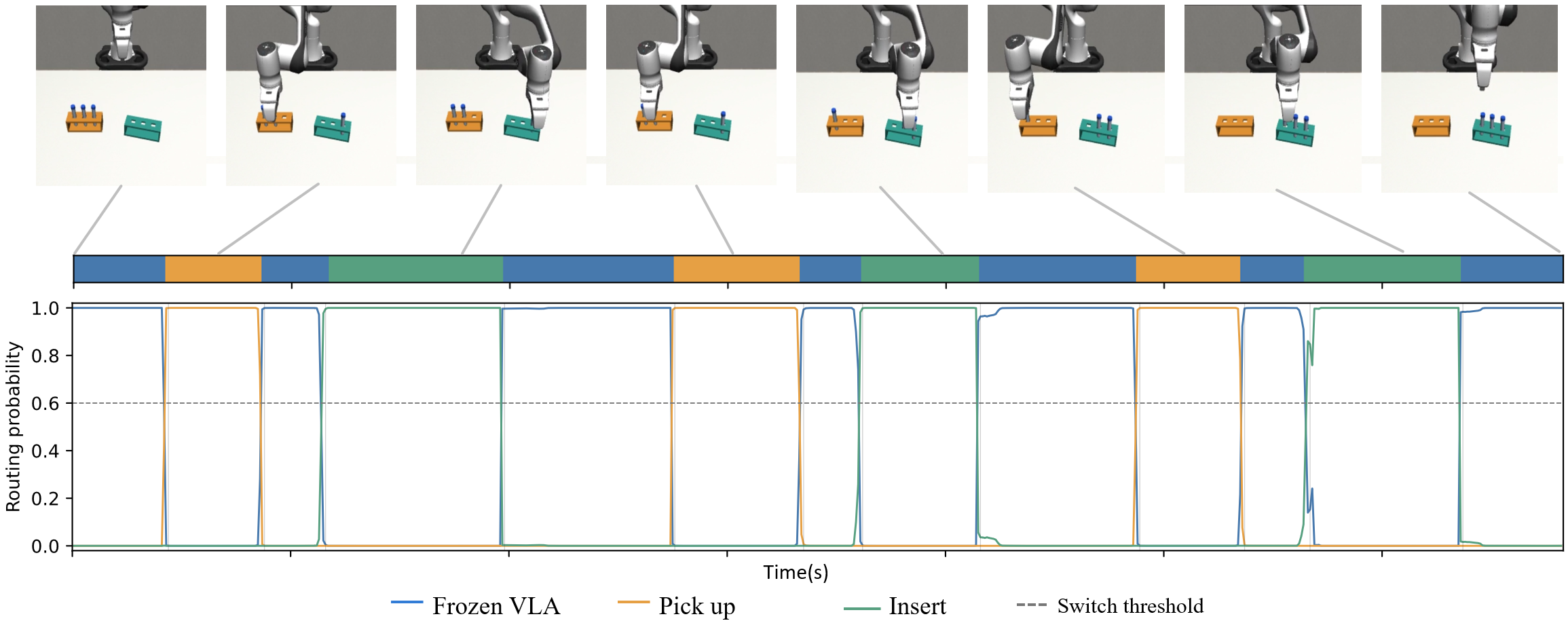}
  \caption{Examples of closed-loop MARS execution in simulation.}
  \label{fig:exp_demo}
\end{figure}

\subsection{Real-Robot Tube Transport Experiments (Q4)}
\label{sec:real_robot}

Although the preceding experiments demonstrate the effectiveness of MARS in simulation, real-world manipulation introduces additional challenges, including perception errors, control inaccuracies, and contact uncertainties. To evaluate its performance under these conditions, we conduct real-robot experiments on the AgileX piper platform described in Section~\ref{sec:sim_tasks}, following the same Tube Transport task procedure.

As shown in Table~\ref{tab:real_robot_main}, MARS achieves a full-task success rate of 74.00\%, exceeding Frozen VLA by 62.00 percentage points. The sample completion rate also increases from 27.33\% to 83.33\%, with higher success ratios observed in both grasping and insertion. These results indicate that the integration of specialized atomic skills and dynamic routing improves both task completion and execution efficiency in the tested physical environment.

Together with the simulation results in Q3, these findings suggest that MARS can maintain its multi-policy coordination capabilities when deployed on a real robotic platform, supporting its applicability to physical multi-stage manipulation.

\section{Conclusion}
\label{sec:conclusion}
In this paper, we presented MARS, a dynamic atomic-skill routing framework that coordinates a generalist vision-language-action (VLA) policy with task-specific reinforcement learning-based policies for multi-stage robotic manipulation. MARS separates atomic skill optimization from high-level policy selection by training specialized skill policies using shared RL Token representations and VLA reference actions, while integrating temporal context modeling and skill-semantic queries for dynamic policy routing. Experiments on LIBERO and Tube Transport demonstrate the effectiveness of this approach. MARS achieves an average Macro-F1 of 90.47\% across the four LIBERO task suites and improves full-task success rates by 34.00 and 65.63 percentage points over the frozen VLA on LIBERO-Long and Tube Transport, respectively. Ablation studies further support the contributions of temporal modeling and skill-semantic queries, while real-robot experiments demonstrate the feasibility of deploying MARS for physical multi-stage manipulation. These results suggest that combining specialized atomic skills with dynamic policy routing provides an effective approach to coordinating multiple policies in complex robotic manipulation tasks.

Nevertheless, MARS relies on supervised policy labels from existing trajectories to train the router, which may limit its adaptability to new tasks and atomic skills. Future work will explore feedback-driven routing optimization to reduce dependence on manually annotated task stages and further investigate the reliability and generalization of MARS across diverse real-world manipulation scenarios.

\nocite{xu2026rlt,gu2023mamba,liu2023libero}
\bibliographystyle{IEEEtran}
\bibliography{references}
\end{document}